\documentclass[letterpaper]{article} 
\usepackage{aaai25}  
\usepackage{times}  
\usepackage{helvet}  
\usepackage{courier}  
\usepackage[hyphens]{url}  
\usepackage{graphicx} 
\usepackage{natbib}  
\usepackage{caption} 
\usepackage{multirow}
\usepackage{graphicx}
\newcommand{\sys}{ObjVariantEnsemble}

\usepackage{mathtools}
\usepackage{makecell}

\usepackage{subcaption}
\usepackage{xspace}
\def\etc{\textit{etc.}\xspace}
\def\eg{\textit{e.g.}\xspace}
\def\ie{\textit{i.e.}\xspace}

\usepackage{xcolor}

\usepackage{algorithm}
\usepackage{algorithmic}

\usepackage{newfloat}
\usepackage{listings}
\DeclareCaptionStyle{ruled}{labelfont=normalfont,labelsep=colon,strut=off} 
\floatstyle{ruled}
\newfloat{listing}{tb}{lst}{}
\floatname{listing}{Listing}
\title{ObjVariantEnsemble: Advancing Point Cloud LLM Evaluation in \\ Challenging Scenes with Subtly Distinguished Objects}
\author{
    Qihang Cao\textsuperscript{\rm 1,\rm 2},
    Huangxun Chen\textsuperscript{\rm 2}\thanks{The work was done during Qihang Cao's research internship at HKUST(GZ). Corresponding author: Huangxun Chen. }
}
\affiliations{
    \textsuperscript{\rm 1}Shanghai Jiao Tong University, Shanghai, China \\
    \textsuperscript{\rm 2}Hong Kong University of Science and Technology (Guangzhou), Guangzhou, China 

    john\_cao@sjtu.edu.cn, huangxunchen@hkust-gz.edu.cn
}

\usepackage{bibentry}

\begin{document}

\maketitle

\begin{abstract}
3D scene understanding is an important task, and there has been a recent surge of research interest in aligning 3D representations of point clouds with text to empower embodied AI. However, due to the lack of comprehensive 3D benchmarks, the capabilities of 3D models in real-world scenes, particularly those that are challenging with subtly distinguished objects, remain insufficiently investigated. To facilitate a more thorough evaluation of 3D models' capabilities, we propose a scheme, \sys, to systematically introduce more scenes with specified object classes, colors, shapes, quantities, and spatial relationships to meet model evaluation needs. More importantly, we intentionally construct scenes with similar objects to a certain degree and design an LLM-VLM-cooperated annotator to capture key distinctions as annotations. The resultant benchmark can better challenge 3D models, reveal their shortcomings in understanding, and potentially aid in the further development of 3D models. 
\begin{links}
\link{Project Page}{https://ove-benchmark.github.io/OVE/}
\end{links}
\end{abstract}

%

\section{Introduction}

\begin{figure}[t]      
    \centering
    \includegraphics[width=0.85\linewidth]{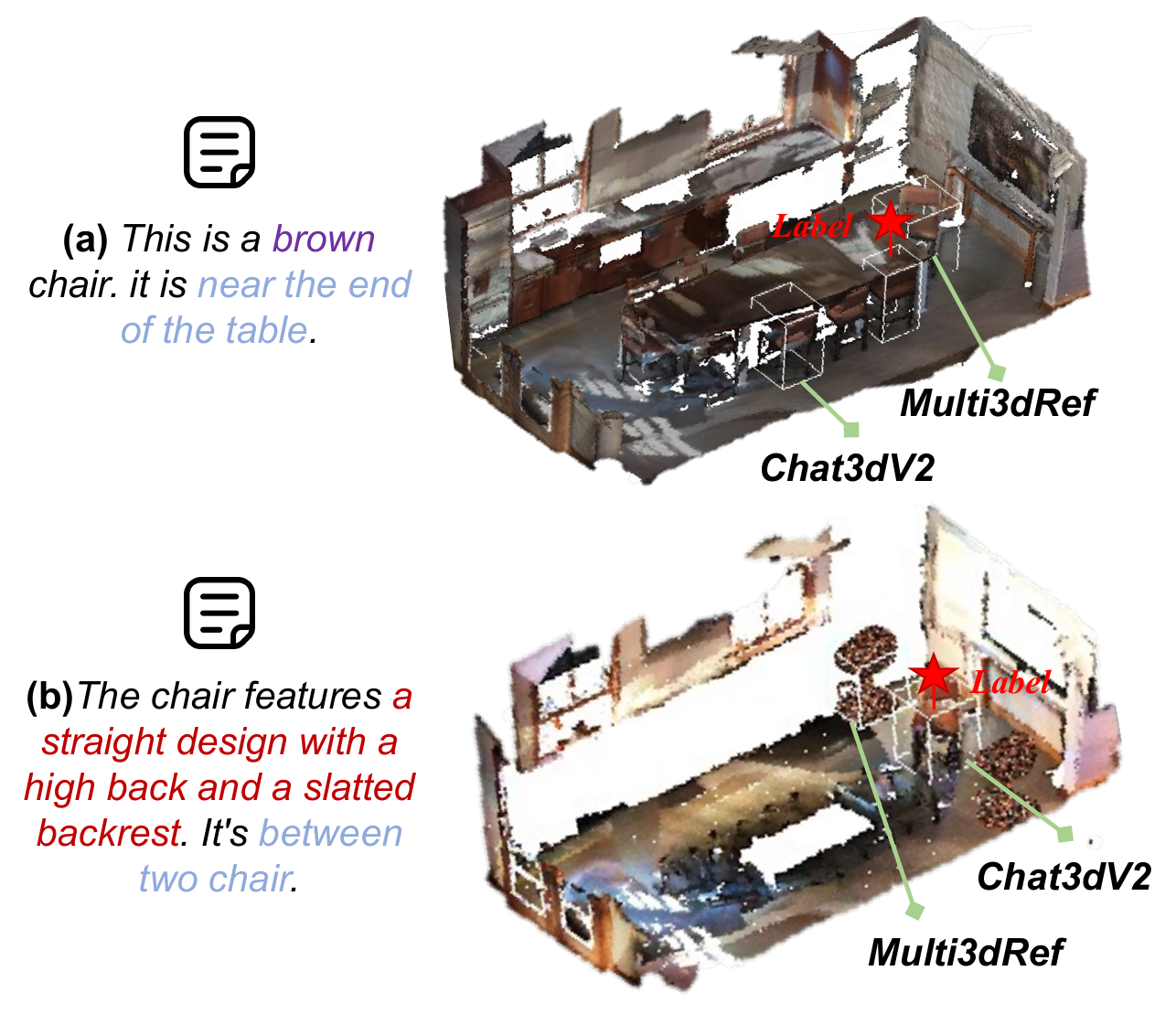} 
    \caption{Comparison of 3D grounding benchmarks in challenging scenes: (a) one scene in ScanNet/ScanRef where the text is insufficient to accurately locate a chair. (b) one scene in \sys~where one model accurately identifies targets with sufficient descriptions.} 
    \label{fig:motivate}
\end{figure}

Building machine perception that can understand our 3D world has been an attractive pursuit in recent years. Prior works~\cite{chang2015shapenet,uy2019revisiting,yu2022point,sun2022benchmarking} have focused on learning 3D representations from point clouds, making significant progress in object classification. Due to the success of large language models (LLMs)\cite{touvron2023llama,liu2024visual}, interest has expanded beyond traditional object classification tasks. Recent works, such as PointLLM\cite{xu2023pointllm} and 3D-LLM~\cite{hong20233d}, aim to align the latent representations of textual descriptions with 3D point clouds, allowing machine perception systems to interpret and interact with the physical world more effectively through text-based instructions. For instance, as illustrated in Figure~\ref{fig:motivate}, if a system can accurately identify a target in a scene based on text descriptions, it could significantly enhance the intelligence of robotics applications, enabling tasks such as completing household chores through verbal instructions~\cite{li2023behavior,ge2024behavior} or improving human-robot collaboration~\cite{xiao2022robotic,team2023human,huang2023voxposer}.

However, the path towards the aforementioned vision of 3D scene understanding is currently hindered by the lack of comprehensive 3D benchmarks. Existing scene-level benchmarks are primarily built on ScanNet~\cite{dai2017scannet} that consists of 1600+ scans of real scenarios. Subsequent works, including ScanRef~\cite{chen2020scanrefer}, Multi3DRef~\cite{zhang2023multi3drefer}, ScanQA~\cite{azuma2022scanqa}, \etc continue to refine and enhance the text annotations to ScanNet, aiming to facilitate 3D model evaluation and development.

\begin{table}[t]
\centering
\renewcommand{\arraystretch}{1.5} 
\begin{tabular}{>{\centering\arraybackslash}p{2cm}|>{\centering\arraybackslash}p{0.6cm}|>{\centering\arraybackslash}p{1.5cm}|>{\centering\arraybackslash}p{2.6cm}}
\Xhline{1pt}
\textbf{3D Benchmark} & \textbf{\#Obj} & \textbf{s-t pairs (wd/wod)} &\textbf{Anno (\#Obj)}                       \\ \Xhline{0.7pt}
ModelNet              & 12k              & -                                 & \multirow{4}{*}{Cl}                         \\
ShapeNet              & 51k              & -                                 &                                                \\
ScanObjectNN          & 15k              & -                                 &                                                \\
OmniObject            & 6k               & -                                 &                                                \\ \hline
ScanRefer             & -                & 39k/12k                           & Cl+s/co/l (1)                         \\
Multi3DRef            & -                & 33k/29k                           & Cl+s/co/l (0/1/2..)            \\
\textbf{OVE(Ours)}    & -                & \textbf{75k/12k}                  & Cl+s/co/l (1 wd)\\ 
\Xhline{1pt}
\end{tabular}
\caption{Benchmark Comparison: “s-t pairs (wd/wod)” represent scene-text pairs with and without distractors. “Anno(\#Obj)” refers to the types of annotation information available for the target object. “Cl,” “s,” “co,” and “l” stand for Class, shape, color, and location, respectively. Numbers in parentheses indicate the number of targets the annotation can locate within the scene. 
Compared to other benchmarks, our dataset offers more challenging scene-text pairs.}
\label{tab:comparison}
\end{table}

Despite many efforts, 3D benchmarks still fall short of keeping pace with the needs of 3D model development. 

\noindent\textbf{(i) Small Scale}. Compared to VLM works~\citep{radford2021learning,li2022blip,liu2024visual}, which require large-scale text-image pairs, 3D benchmarks, especially those from real-world scans, are significantly smaller in scale.

\noindent\textbf{(ii) Insufficient Annotation}. Beyond data scale, the annotation granularity is more crucial in determining whether we can objectively evaluate 3D model strengths and weaknesses for potential improvement. Figure~\ref{fig:motivate}(a) demonstrates a scene and its associated annotation from ScanRef. Though the predictions of two 3D models, Multi3DRef and Chat-3D-v2~\cite{huang2023chat} do not align with ground truth label, it is hard to say the models are incapable, since the text itself contains certain ambiguities and can refer to multiple objects in the scene.

\noindent\textbf{(iii) Lack of customizable challenge levels}. Current 3D benchmarks are based on limited scene layouts and in-scene object combinations and placements. While they may include certain challenging scenes, they are not customizable, which can potentially lead to overfitting. It is better to have diverse scene data with customizable difficult level to facilitate sufficient evaluation of model capabilities.

Motivated by above gaps and persistent needs for more comprehensive 3D datasets, 
\textbf{we propose to synergize object-level and scene-level 3D point cloud resources to construct scenes with greater variety}. 
The community has developed comprehensive object-level 3D resources, \eg, ModelNet~\cite{sun2022benchmarking}, ShapeNet~\cite{chang2015shapenet}, ScanObjectNN~\cite{uy2019revisiting}, and OmniObject~\cite{wu2023omniobject3d}, among others as shown in Table~\ref{tab:comparison}. These object-level dataset generally offer broader object classes and rich variants within a class compared to scene-level ones, as detailed our project page. This makes them highly promising as a candidate pool to ensemble new scenes, especially challenging ones featuring subtly distinguished objects. For instance, Figure~\ref{fig:motivate}(b) showcases an ensembled scene with 3 chairs, one from ScanRef and the others retrieved from OmniObject. 
Besides scene construction, \textbf{we further integrate LLMs and 2D Vision-Language Models (VLMs) together to build an automated and fine-grained annotation pipeline}. Specifically, we instruct the LLM to guide the VLM in focusing on the differences between target and distractors, summarizing their distinctions in terms of multiple aspects (color/shape/location/...) as the annotation, \eg, the text in Figure~\ref{fig:motivate}(b).

\begin{figure}[t]      
    \centering
    \includegraphics[width=0.9\linewidth]{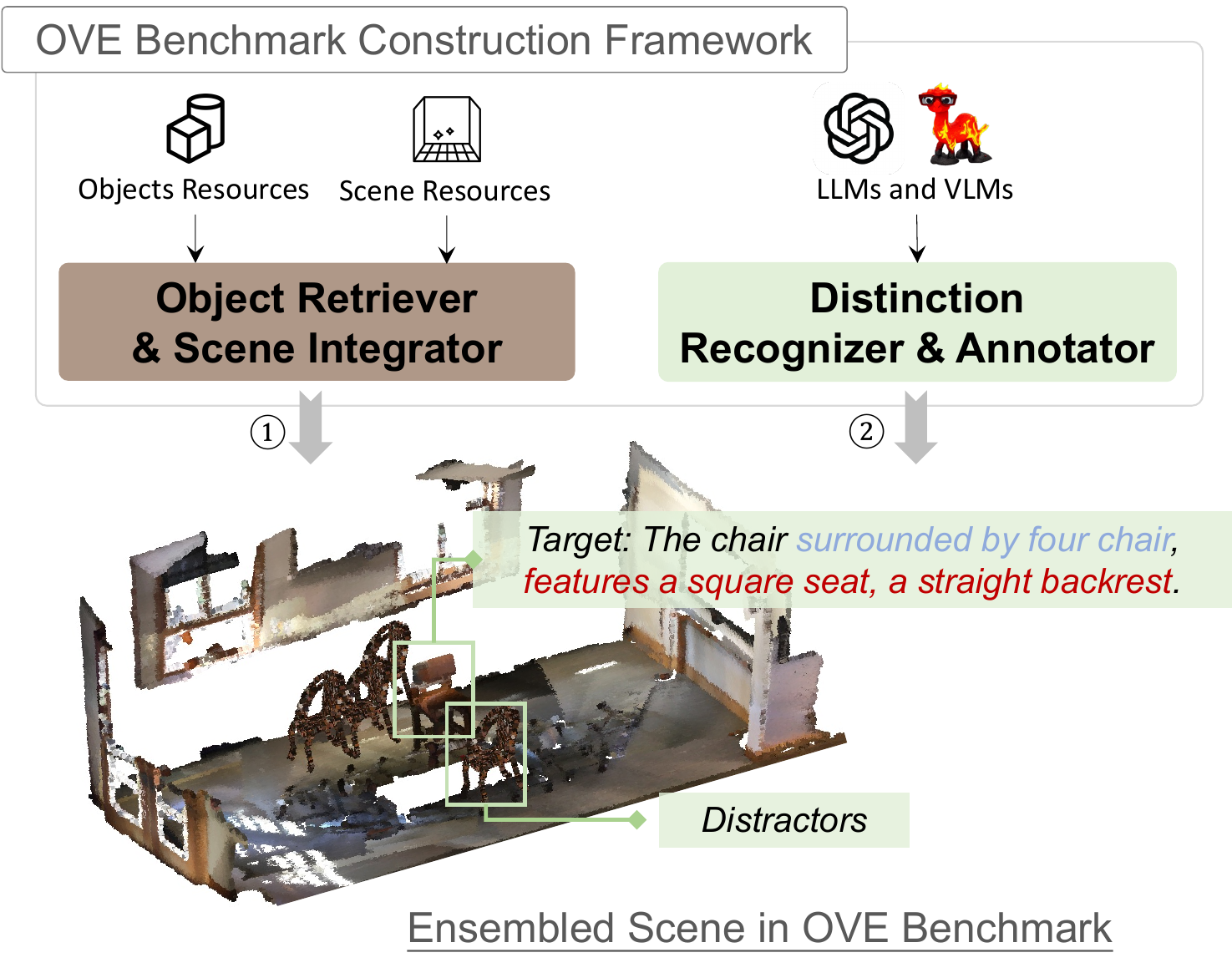} 
    \caption{OVE Benchmark Construction Overview.} 
    \label{fig:overview}
\end{figure}

Our technical framework, \textbf{O}bj\textbf{V}ariant\textbf{E}nsemble (OVE) is shown in Figure~\ref{fig:overview}. We will elaborate the technical details in \$\ref{sec:design}. In summary, we make the following contributions:

\noindent$\bullet$ We designed an effective 3D scene construction and annotation framework to significantly expand the 3D dataset, resulting in 75k newly created scenes and associated fine-grained annotations, as summarized in Table~\ref{tab:comparison}.

\noindent$\bullet$ We developed a systematic and flexible method to introduce subtly distinguished objects adjacent to the target, along with an automated scheme to capture their key differences. This enriches the semantic complexity of the benchmark, enabling a more in-depth evaluation of 3D model. 

\noindent$\bullet$ We evaluate state-of-the-art 3D understanding models on OVE benchmark and provide a fine-grained analysis to reveal the limitations of existing models in pure spatial relationship reasoning without visual features like shape, guiding potential directions for model improvement.

\begin{figure*}[t]      
    \centering
    \includegraphics[width=0.85\textwidth]{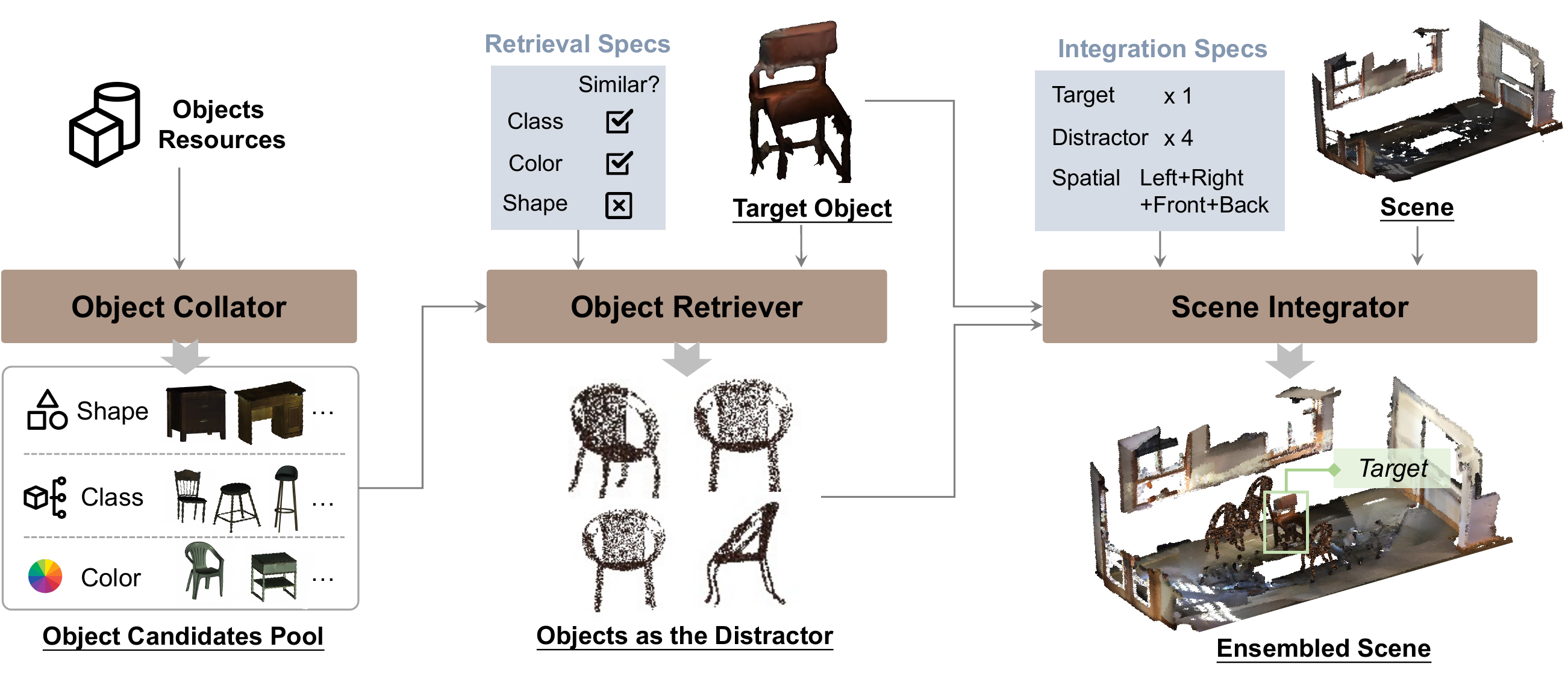}  
	\caption{\sys~Scene Data Generation Framework. (For clearer illustration here, we only plot the target and distractors in the scene, without other background objects.)}
	\label{fig:scene_integrate}
\end{figure*}

\section{Related Works}
In this paper, we focus on 3D understanding with point cloud data. We briefly discuss related works as follows.

\subsubsection{3D Point Cloud Benchmarks}
Existing 3D point cloud datasets fall into two categories: object-level and scene-level. 
Object-level datasets~\citep{chang2015shapenet, uy2019revisiting, sun2022benchmarking, wu2023omniobject3d} feature individual objects in various classes and styles, along with class-level text, to facilitate the evaluation of downstream tasks such as classification and object partial segmentation.
Scene-level datasets are primarily built on ScanNet, which mainly covers indoor scenes scanned from the real world. An important subsequent work, ScanRefer, annotates objects with natural language descriptions to support the evaluation of 3D grounding tasks.
Additionally, ReferIt3D \citep{achlioptas2020referit3d} has introduced two datasets: Sr3D, with textual annotations generated based on predefined templates, and Nr3D, with human-annotated descriptions for more fine-grained understanding. Multi3DRef generalizes ScanRef from single-target grounding to various target grounding, \ie, 0, 1, 2, and so on. However, these benchmarks are inherently limited by the available scenes in ScanNet, which are fixed and limited. Our work aims to break this constraint by incorporating additional object resources to assemble more scenes, particularly challenging ones.
It is worth mentioning that while Multi3DRef handles scenarios where one text corresponds to varying numbers of objects, our approach focuses on scenarios where one text corresponds to a single object with multiple distractors.

\subsubsection{Point Cloud LLM}
Owing to the great success of LLMs, there has been increasing interest in aligning various modalities, including point clouds with textual data, to apply the common-sense knowledge learned by LLMs to multi-modal understanding \citep{han2023imagebind}.
Technically, this demands a robust point cloud feature extractor, which has driven many prior works on 3D representation learning \citep{yu2022point, huang2023clip2point, zeng2023clip2, hong20233d}. It is noteworthy that many of these works leverage relatively mature VLMs driven by large-scale image-text pairs to help learn representations of 3D point clouds. Recently, the release of large-scale point cloud datasets \citep{wu2023omniobject3d, deitke2023objaverse} has enabled scalable pretraining \citep{xue2023ulip, liu2024openshape, zhang2023uni3d}, significantly advancing the capabilities of point cloud encoders.

However, when dealing with scene-level point clouds, object-level encoders often struggle to manage complex spatial relationships. 3D-LLM attempts to address this by projecting 3D features into 2D to integrate them into LLMs and incorporating positional embeddings and learnable position tokens. Meanwhile, Chat-3D-v2 \citep{huang2023chat} segments scenes into object-level point clouds and integrates the features and positional information obtained from object-level 3D encoders. Additionally, Any2Point \citep{tang2024any2point} has proposed positional encoding that merges spatial features across 3D, 2D, and 1D, mitigating alignment errors caused by spatial relationships.
Despite these design efforts, existing models still have limited pure spatial relationship reasoning over point cloud data, as evidenced by our evaluation in \S~\ref{sec:eval}.

\section{Benchmark Data Construction}
\label{sec:design}

In this section, we will provide an overview of the OVE framework and then illustrate the technical details.

\begin{figure*}[t]      
	\centering
    \includegraphics[width=0.9\textwidth]{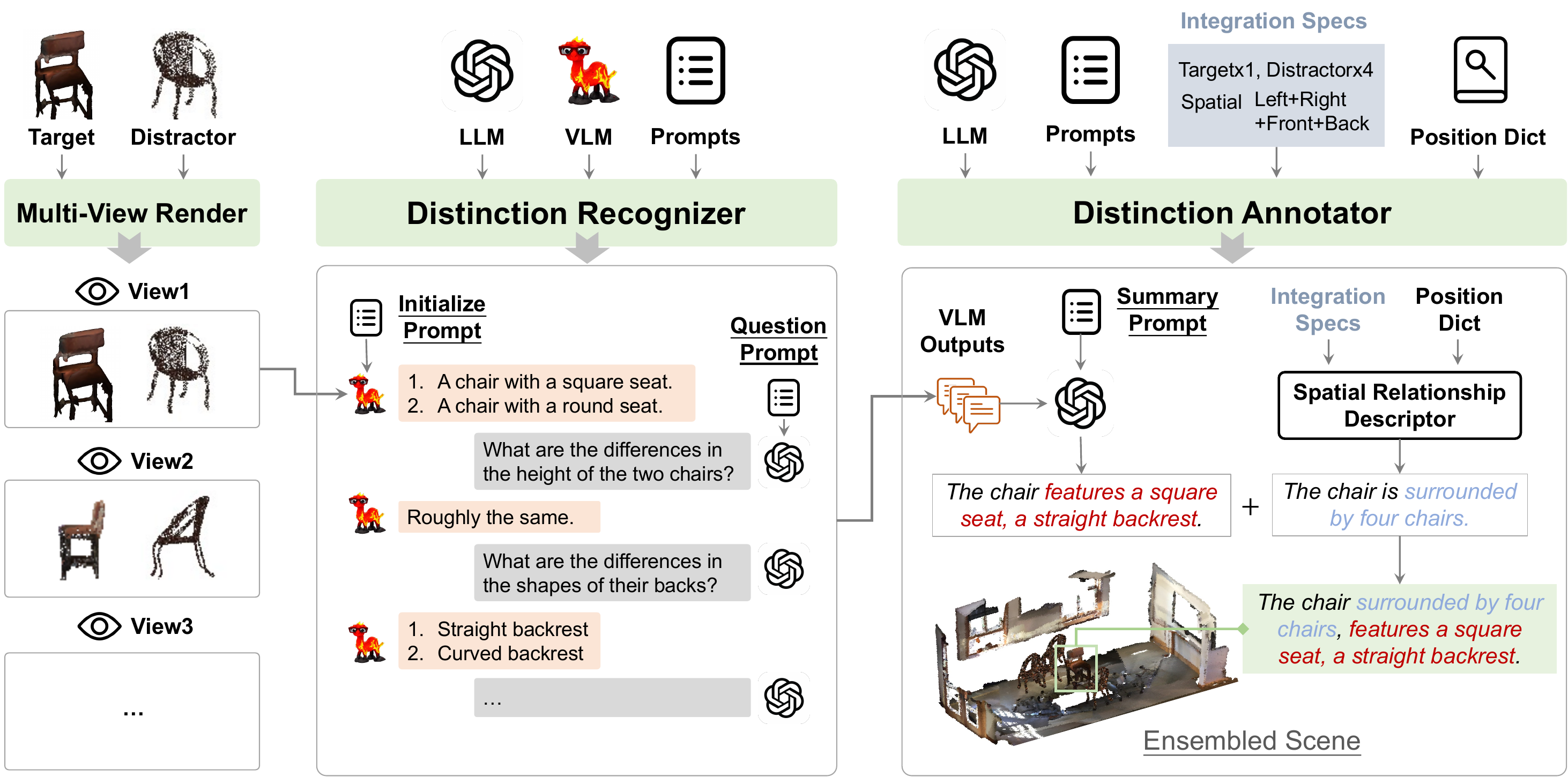} 
	\caption{Process for Capturing Annotations with Key Distinguishing Information. We render multi-view images and use LLAVA\citep{liu2024visual} to extract differences from various perspectives. A LLM is then employed to generate questions based on previous Q\&A interactions. Finally, we use LLM to summarize the key differences from all descriptions.} 
	\label{fig:distinct_anno}
\end{figure*}

\subsection{OVE Overview}

As shown in Figure~\ref{fig:overview}, OVE consists of two main modules: 

\noindent\textbf{Object Retriever \& Scene Integrator}: This module sorts available object-level 3D resources by their features, including class, color and shape, and then assembles multiple objects with varying levels of similarity into a specific scene background to construct a new scene.

\noindent\textbf{Distinction Recognizer \& Annotator}: Most object-level datasets only have class labels, as indicated in Table~\ref{tab:comparison}, which are insufficient as effective annotations in the context of our constructed scene. Therefore, this module leverages the capabilities of advanced LLMs and VLMs to extract the key characteristics that distinguish the target object from others and summarize these as annotations. 

In the following, we illustrate these two modules in detail.

\subsection{Object Retriever \& Scene Integrator}
Figure~\ref{fig:scene_integrate} shows the detailed procedure for assembling a new scene by fully leveraging object-level and scene-level 3D resources. 
The construction process involves three steps: object collator, object retriever and scene integrator. 

\subsubsection{Object Collator} 
As shown in Figure~\ref{fig:scene_integrate}, this step aims to prepare an object candidates pool for the subsequent retrieval stage. To achieve this, we sort the object-level resources based on class, color, and shape. Specifically, we retain the object class from its original source dataset. We then represent color using the standard 3-tuple RGB values and use the L2-norm to quantify color similarity. Shape is harder to quantify, so to reduce potential bias, 
we currently focus on standard shapes, mainly including cuboids, L-shapes and spheres that are prevalent in our candidates to facilitate subsequent scene construction with similar-shaped objects.


\subsubsection{Object Retriever}
This step aims to retrieve objects based on retrieval specifications. A retrieval specification consists of: i) a target object, where the retrieved objects are expected to preserve certain similarities with the target one, and ii) similarity dimension, where the distinction could occur in class, color, shape, or a combination of these factors.
The example in Figure~\ref{fig:scene_integrate} shows a case of finding an object with the same class and color but a different shape compared to the target, a brown chair. The object retriever searches over the object candidate pool and identifies another brown chair with a different shape to serve as the distractor. 
When choosing distractors, we prioritize real-scanned objects (OmniObject /ScanObjectNN). If no suitable objects found, we then consider CAD data (ModelNet/ShapeNet).

\subsubsection{Scene Integrator} This step involves seamlessly integrating the retrieved distractors and the target object into a real-world scanned background extracted from the scene-level dataset. As shown in Figure~\ref{fig:scene_integrate}, the integration specifications determine the number of distractors, which adjusts the challenge level of the assembled scene for model understanding. With more distractors, it generally becomes more difficult for the model to locate the correct target object.

Additionally, we specify the spatial relationship between the target and distractors. 
To better cover various possible object placements, we use six spatial primitives: \texttt{left}, \texttt{right}, \texttt{front}, \texttt{back}, \texttt{up} and \texttt{down}. Building upon these primitives and their combinations, we define a total of 13 spatial predicates (detailed in our project page), including terms such as \texttt{lower left}, \texttt{between}, \texttt{surrounding} for scenarios with multiple distractors.

It is worth mentioning that we carefully ensure the coherence of the ensembled scene. First, the target object is segmented from the scene-level dataset, and we use its original scene background as the base for constructing the new scene. 
Second, for the retrieved distractor objects, we properly resample, rescale, and reorient them to ensure coherence with the target and the scene, effectively mitigating scale mismatches between their original datasets and the target’s ones.
Third, after enforcing spatial relationship between distractor and target object, we calculate their bounding boxes to only exclude overlapping background objects to ensure scene reasonability. We manually screen assembled scenes and have not identified noticeably odd aspects.

The output of Object Retriever\&Scene Integrator is newly ensembled scenes, each with a target object and a few distractor objects that share certain similarities with the target, as illustrated in the lower right of Figure~\ref{fig:scene_integrate}.

\subsection{Distinction Recognizer \& Annotator}

3D scenes alone cannot serve to evaluate the model without associated annotations. 
Figure~\ref{fig:distinct_anno} shows the detailed procedure to derive the key distinctions between the target object and the distractors to obtain an accurate annotation. 
This process involves two main steps: distinction recognizer and distinction annotator.

\subsubsection{Distinction Recognizer}
This step aims to capture as many distinctions as possible, even subtle ones, between the target and its distractors. We employ two methods to achieve this goal: 
i) \underline{multi-view recognizer}. To avoid missing critical distinctions, we render the target-distractor pair from multiple perspectives and then let VLMs recognize the differences. 
ii) \underline{iterative difference capturing}: we design an iterative QA process between VLM and LLM to enhance the comprehensiveness of distinction recognition. Specifically, after initializing the VLM to recognize differences, we prompt a LLM to continuously ask VLM about new potential distinction dimensions, forcing the VLM to capture more differences, as illustrated in Figure~\ref{fig:distinct_anno}. This QA process would be repeated multiple rounds (6-7 in our case) to ensure sufficient and high-quality distinctions are captured. The detailed algorithm is provided in Algorithm~\ref{algo}, and the involved prompts are detailed in our project page. 

\begin{algorithm}[t]
\caption{Target Annotation Process}
\label{algo}
\begin{algorithmic}[1]
\FOR{each (tgt, distr) in pairs}
    \FOR{each rnd in rounds}
        \FOR{each $v$ in views}
            \STATE $\text{tgt}\_\text{desc}.\text{append}(v: \textbf{LLaVA}(\text{tgt}\_\text{img}))$
            \STATE $\text{distr}\_\text{desc}.\text{append}(v: \textbf{LLaVA}(\text{distr}\_\text{img}))$
            \STATE $\text{img} \gets \textbf{Concat}(\text{tgt}\_\text{img}, \text{distr}\_\text{img})$
            \STATE $\text{cap} \gets \textbf{IterCap}(\text{img},\text{iter}\_\text{rounds})$ 
            \STATE $\text{cap}\_\text{all}.\text{append}(v:\text{cap})$
        \ENDFOR
        \STATE $\text{sum} \gets \textbf{GPT}(\text{cap}\_\text{all}, \text{SUM\_P\_2})$
        \STATE $\text{tgt}\_\text{sum} \gets \textbf{GPT}(\text{tgt}\_\text{desc}, \text{SUM\_P\_2})$
        \STATE $\text{distr}\_\text{sum} \gets \textbf{GPT}(\text{distr}\_\text{desc}, \text{SUM\_P\_2})$
    \ENDFOR
    \STATE $\text{desc} \gets \textbf{GPT}(\text{tgt}\_\text{sum}, \text{distr}\_\text{sum}, \text{sum}, \text{SUM\_P\_3})$
\ENDFOR
\end{algorithmic}
\end{algorithm}

\subsubsection{Distinction Annotator} All answers from VLM above are then compiled and fed into a summarization LLM to distill the most essential distinction information, as shown in Figure~\ref{fig:distinct_anno}. It is worth mentioning that during the distinction recognition stage, VLMs are primarily used to identify distinctions in terms of detailed shape and color. For a complete annotation, we further enhance the spatial location information based on the integration specifications used for scene construction. 
With this, we can further describe the target's location in the scene and combine this with others to create a fine-grained annotation, as illustrated in Figure~\ref{fig:distinct_anno}.


OVE ensures high-quality annotations through: (i) Basic attribute annotations (e.g., class, color, position) are guaranteed during scene construction. (ii) LLM-VLM collaboration helps capture high-quality visual difference annotations. Multi-round QA is used to reduce hallucinations. In each round, the LLM prompts the VLM to focus on a single attribute (e.g., height). The LLM asks the same question multiple times and discards overly verbose answers. The summary prompt explicitly instructs the LLM to exclude unspecified aspects, such as texture. (iii) We continuously sample annotations for manual verification to ensure quality.

\subsection{OVE Benchmark Summary}

\begin{figure}[t]      
	\centering
        \includegraphics[width=0.85\linewidth]{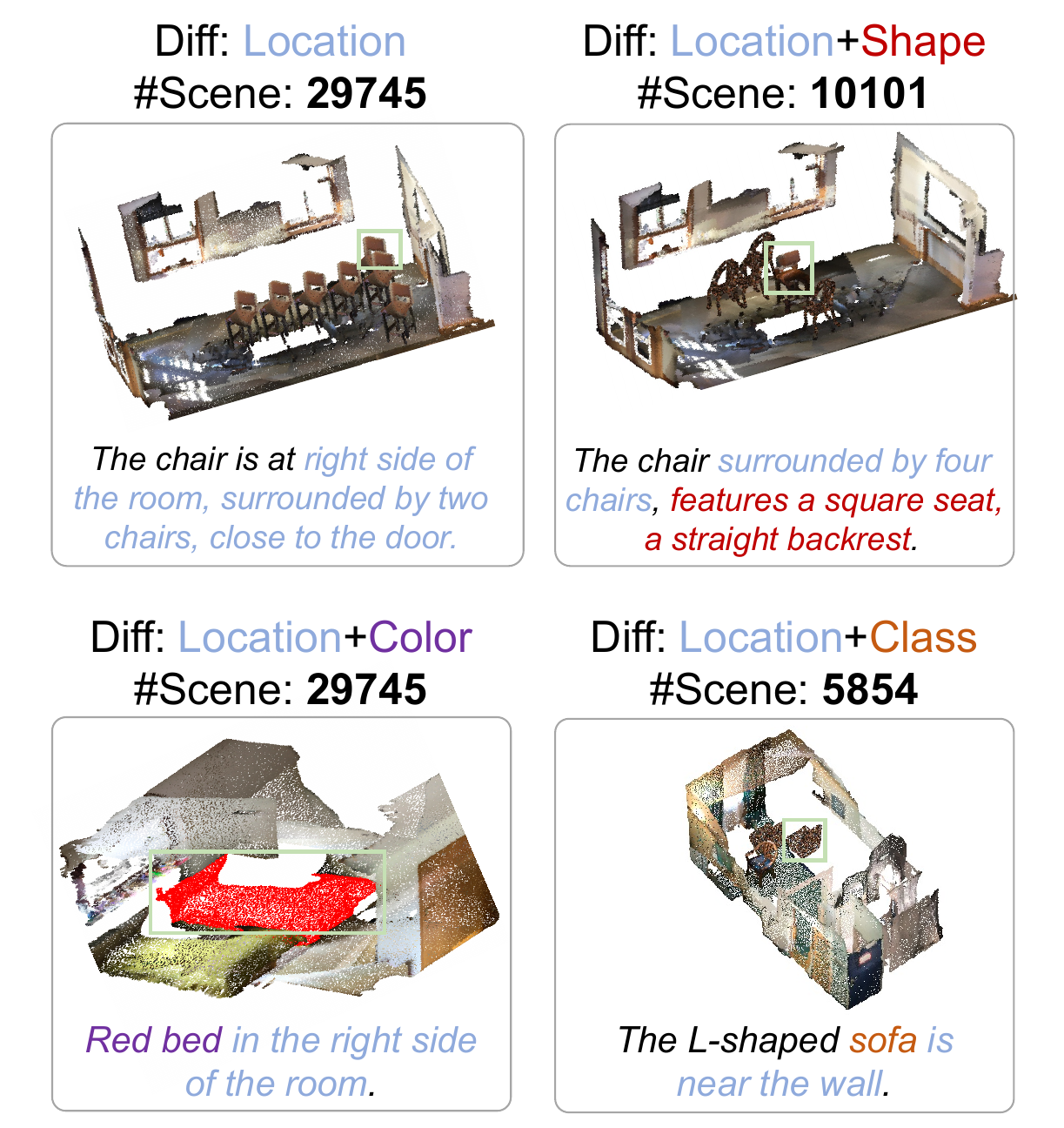} 
	\caption{OVE Benchmark Summary} 
	\label{fig:ove_sum}
\end{figure}

In a nutshell, we construct comprehensive scenes with various challenge levels and distinction dimensions. The resultant benchmark is summarized in Figure~\ref{fig:ove_sum}. We support four different distinction types between the target object and distractors: \texttt{location}, \texttt{location+shape}, \texttt{location+color}, and \texttt{location+class}. Compared to existing scene-level 3D datasets, our work substantially expands the semantic richness of 3D benchmarks. The newly constructed scenes, with customizable challenge levels and fine-grained distinction annotations, can be used to better evaluate and develop 3D models. 
Moreover, our highly customizable pipeline can be seamlessly extended to construct more tasks during scene creation, including 3D object counting, captioning, QA, \etc.

\begin{figure*}[t!]
     \centering
     \begin{subfigure}[b]{0.48\linewidth}
         \centering
         \includegraphics[width=\linewidth]{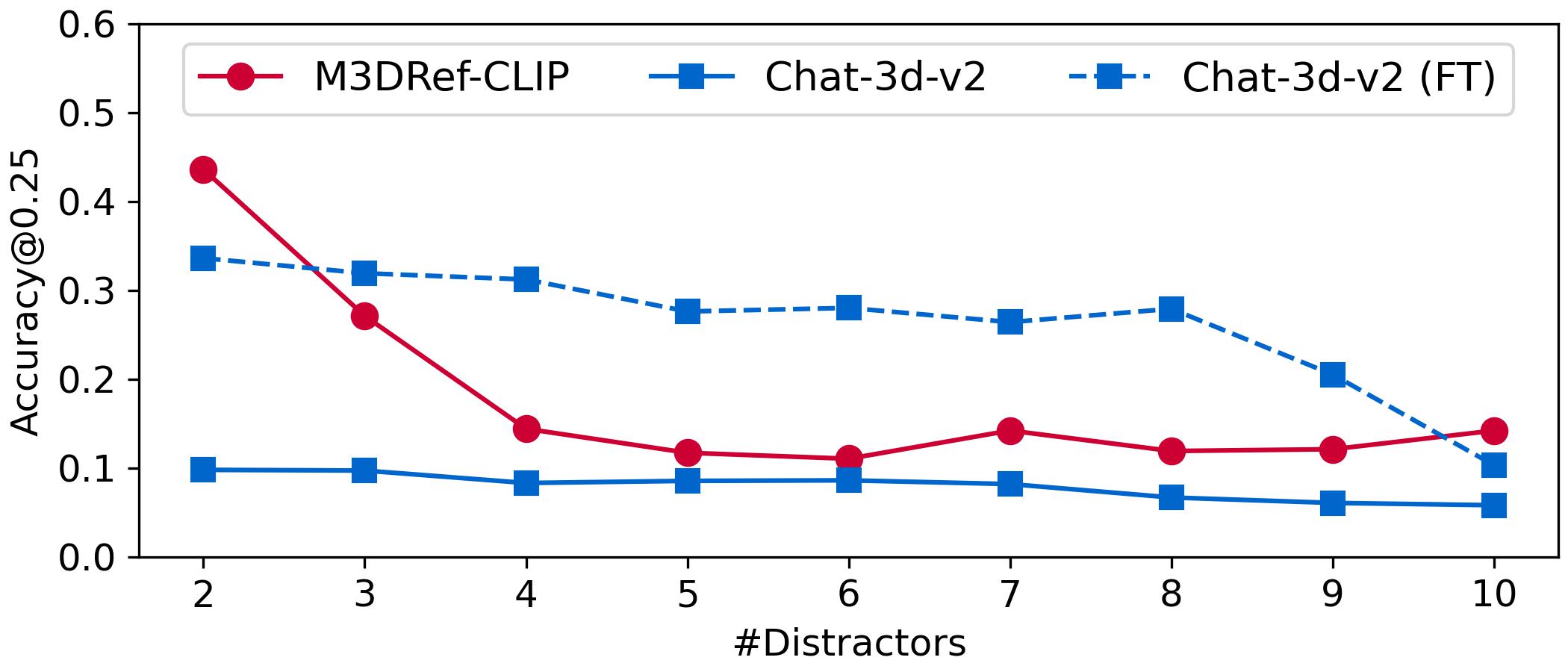}
         \caption{Performance(ACC@0.25) of 3D grounding models on OVE across different numbers of distractors, considering only location.}
         \label{fig:eval_num_distract}
     \end{subfigure}
     \hfill
     \begin{subfigure}[b]{0.48\textwidth}
         \centering
         \includegraphics[width=\textwidth]{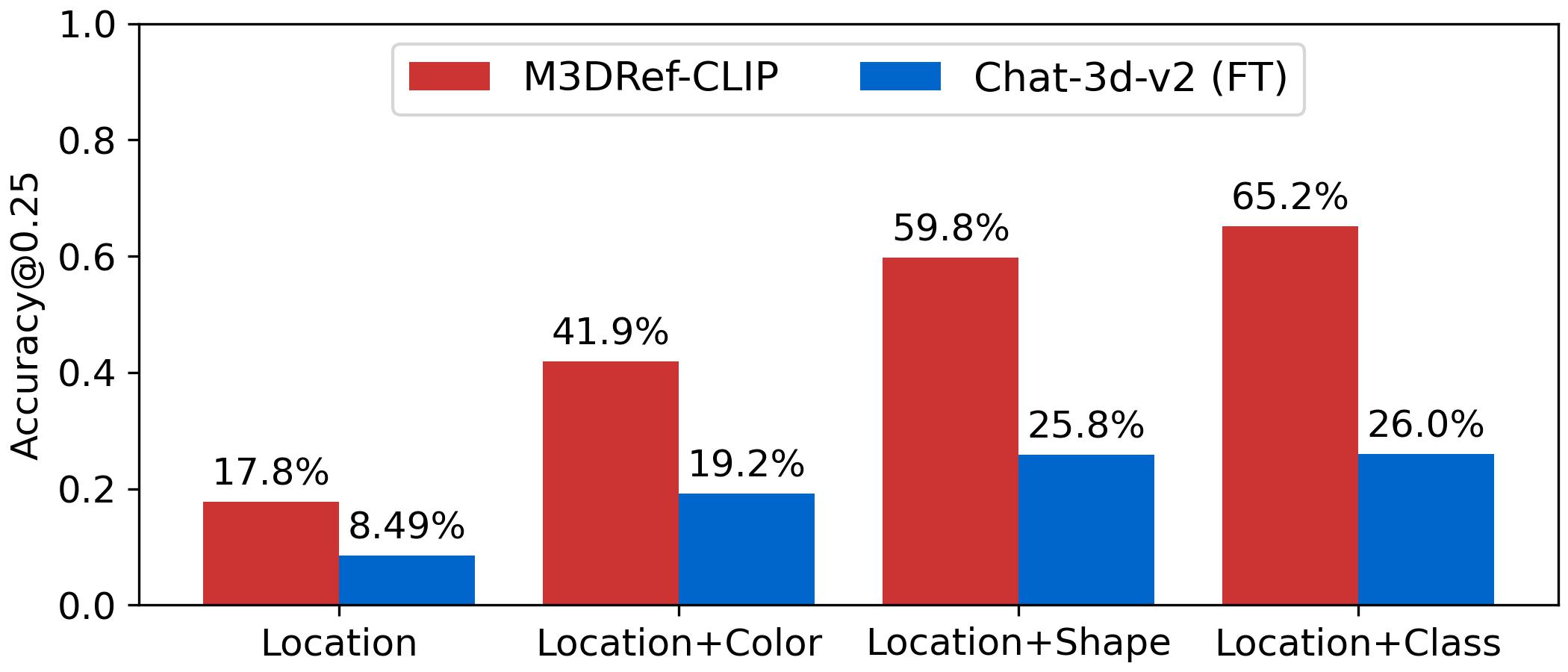}
         \caption{Performance(ACC@0.25) of 3D grounding models on OVE across various distinction types.}
         \label{fig:eval_diff_type}
     \end{subfigure}

     \caption{Experiments conducted on the OVE benchmark.}
\end{figure*}

\section{Benchmark on \sys}
\label{sec:eval}
In this section, we evaluate the performance of state-of-the-art models on our benchmark.

\subsection{Evaluation Setting}

\textbf{Evaluation Task}: 
We adopt 3D grounding as our evaluation task, because it is a fundamental perception task to support higher-level tasks like reasoning~\citep{hong20233d} and planning~\citep{bai2024m3d}.
Specifically, we feed the point cloud scene $S_p$ and its associated textual description $T$ from the OVE benchmark into a model $\mathcal{M}$. The model is then required to predict the target object's bounding box $B_t$ in the scene. Formally, this can be described as follows:

\begin{equation}
    B_t = \mathcal{M}(\mathcal{S}_p, T)
\end{equation}

\noindent\textbf{Evaluation Metric}: 
We adopt Acc@0.25 and Acc@0.5 as evaluation metrics, \ie, prediction accuracy when the intersection over union (IoU) between the predicted and actual bounding boxes reaches 0.25 and 0.5, respectively.

\noindent\textbf{Evaluation Models}:
We chose to evaluate 3D understanding models that rely solely on 3D scene information and 1D textual information to complete 3D grounding tasks. Chat-3D-v2 represents the state-of-the-art performance in 3D grounding, while Multi3DRef takes the initiative in addressing multi-object grounding.
Both models utilize segmentation as an auxiliary task to extract scene features and then employ a fusion module to align 3D features with 1D text. 
Chat-3D-V2 uses Mask3D~\cite{schult2023mask3d} for segmentation and Uni3D~\cite{zhang2023uni3d} for encoding to realize 3D object identification and localization.
We also incorporated a fine-tuned version of Chat-3D-v2 for additional evaluation using Vicuna1.5~\citep{zheng2024judging} and LORA~\citep{hu2021lora}. The training was conducted using a one-stage joint method, with a batch size of 32. 
Multi3DRef employs PointGroup~\cite{jiang2020pointgroup} as its detector and a CLIP-based~\cite{radford2021learning} encoder to realize effective multi-object localization in complex scenes. 
We selected a configured version of M3dRefCLIP that utilizes only the \texttt{use\_color} and \texttt{use\_normal} settings. 
All experiments were performed on 4 NVIDIA A6000 cards.

\begin{figure}[t]      
	\centering
    \includegraphics[width=0.85\linewidth]{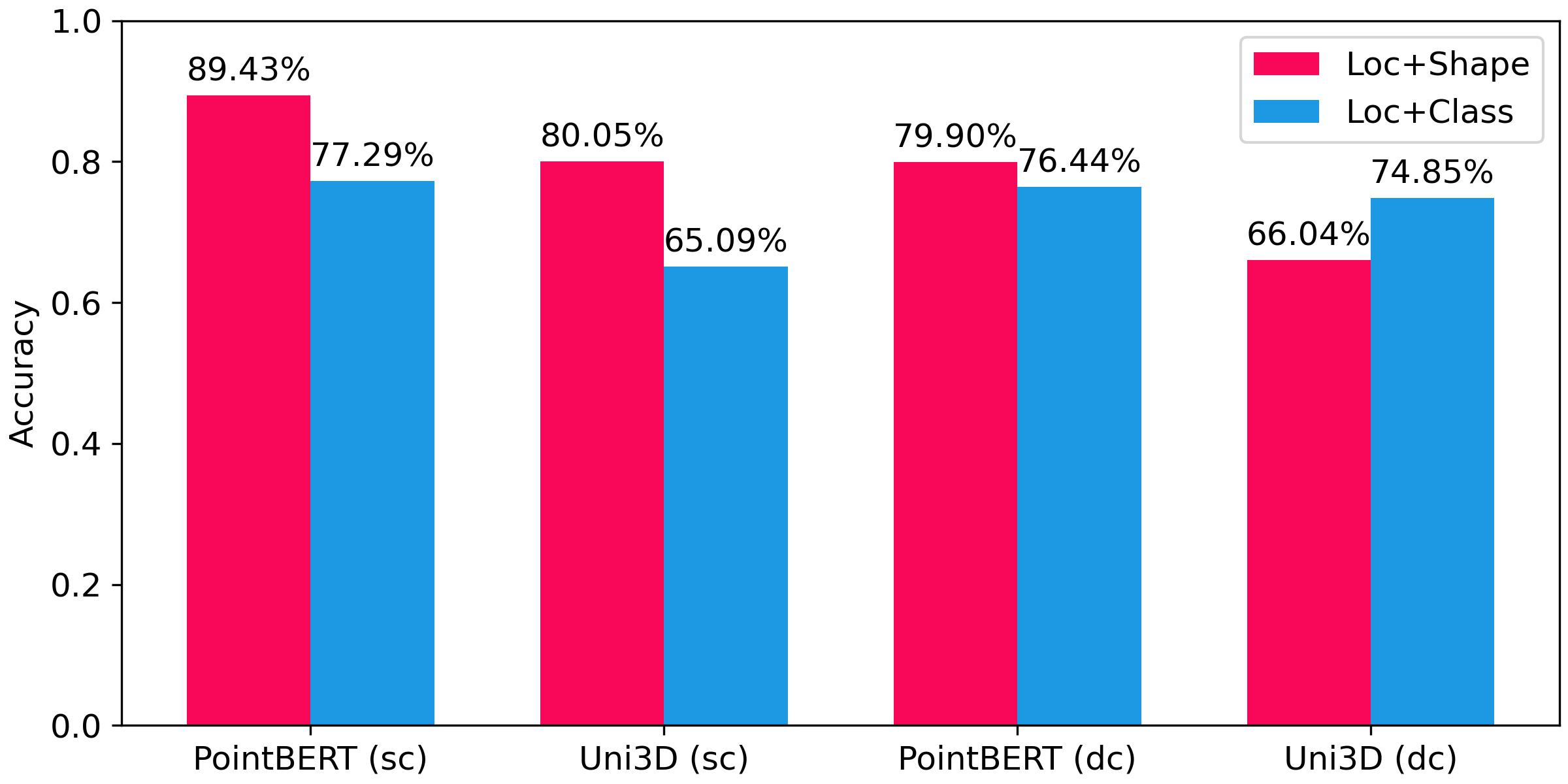} 
	\caption{Performance(mIoU$_\text{I}$(\%)) of 3D representation learning models on segmentation tasks using data resampled by OVE framework.} 
	\label{fig:eval_segment}
\end{figure}
\begin{figure*}[t]     
	\centering
    \includegraphics[width=0.85\linewidth]{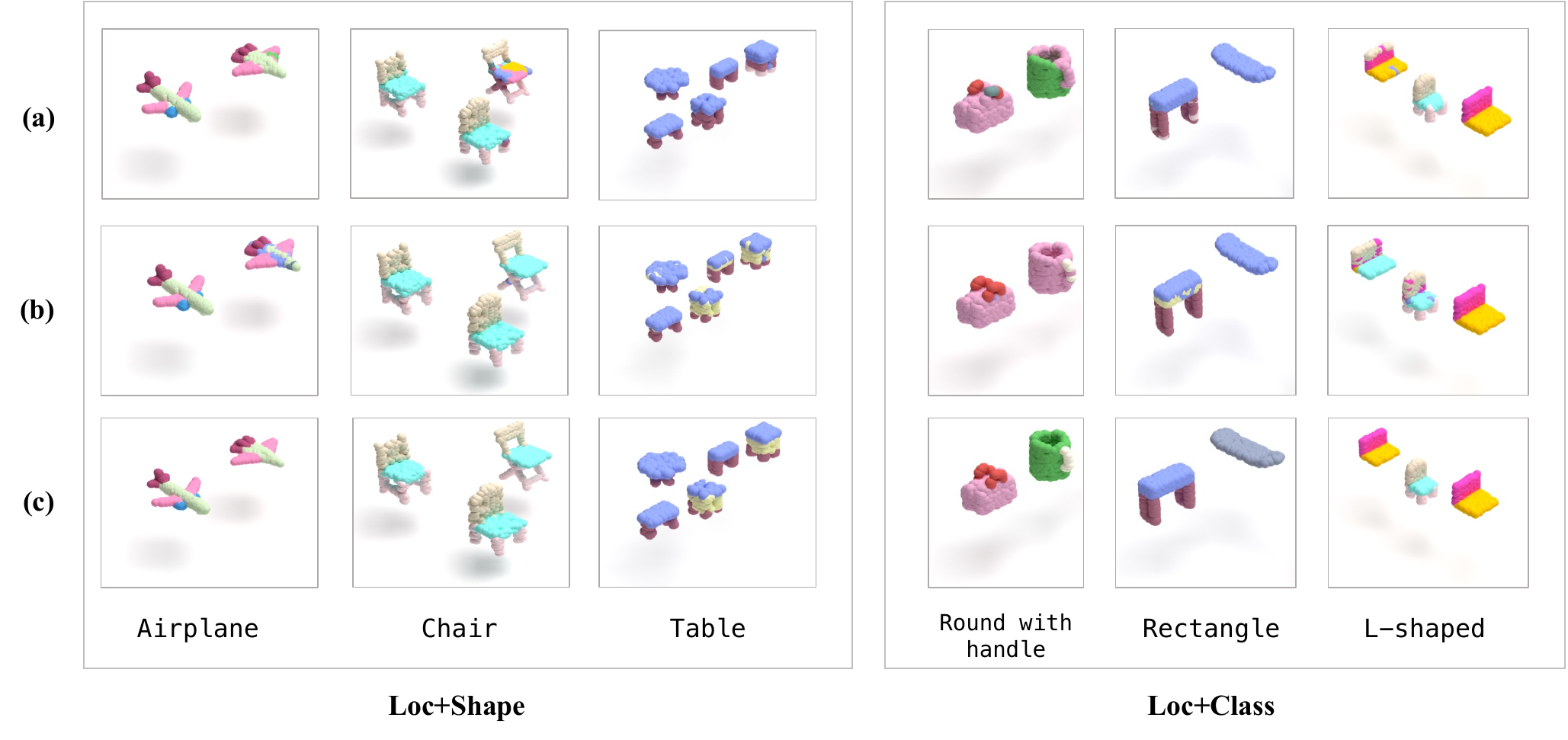}  
	\caption{Visualizations of Segmentation Results on ShapeNet\cite{chang2015shapenet} Resampled: row (a) represents Uni3d, row (b) represents PointBERT, and row (c) represents the ground truth labels.} 
	\label{FIG:segmentation}
\end{figure*}

\subsection{Evaluation Results}
\label{sec-eval-grounding}
\subsubsection{Impact of \#Distractors}

Here, we aim to evaluate whether existing 3D understanding models can maintain their performance, \ie, predict the correct object bounding box, across various challenge levels, specifically with different numbers of distractors. To do this, we selected objects from the validation set of ScanRefer as the target and assembled scenes with the target and varying numbers of distractors, ranging from 2 to 10, to observe the model's performance. The evaluation results are shown in Figure~\ref{fig:eval_num_distract}. It is noted that as the number of distractors increases, the grounding performance of the evaluated models declines to some extent. M3DRef-CLIP achieves the average performance reported in ScanRefer \citep{chen2020scanrefer} with 2 distractors but shows a decline as the number of distracting objects increases. The original Chat-3D-v2 performs rather poorly on the OVE benchmark, significantly less than reported in ScanRefer, potentially due to its overfitting to ScanRefer dataset. However, the fine-tuned version of Chat-3D-v2, using a portion of our dataset for fine-tuning, demonstrates enhanced performance. Compared to ScanRefer, the OVE benchmark supports a more fine-grained model evaluation.

\subsubsection{Impact of Distinction Type}
We further investigate which aspects of existing 3D models' capabilities have the most weaknesses. We used four distinction types with a well-balanced distribution of distractor numbers, as shown in Figure~\ref{fig:ove_sum} to conduct a more fine-grained evaluation on M3DRef-CLIP and the fine-tuned version of Chat-3D-v2. The evaluation results are presented in Figure~\ref{fig:eval_diff_type}.
The results indicate that the current models' ability to align textual information with pure location information is far inferior to their ability to correspond to visual features like shape, color, and class. Even though Chat-3D-v2 includes an additional encoding scheme for location information, its performance does not surpass that of Multi3DRef in challenging scenarios. This suggests that the spatial reasoning capabilities of current 3D understanding models are still ineffective.

\subsubsection{Key Takeaways}
Through the lens of OVE, we reexamine the grounding capability of state-of-the-art 3D understanding models. We highlight how different types of information help distinguish multiple similar objects, revealing the models' strengths and weaknesses.
From Figure~\ref{fig:eval_diff_type}, we observe that shape and class contribute significantly more than color or location, with location playing the least role in object grounding. This raises our natural questions about the effectiveness of model designs related to position embedding. We may need more effort to better include spatial information in 3D representations.

\subsection{Reexamining 3D Representation Learning}
\label{sec-eval-segment}
The performance gain from class/shape information shown in Figure~\ref{fig:eval_diff_type} suggests a hypothesis that the 3D encoders underlying 3D models could be robust, potentially owing to large-scale pre-training on object-level point clouds.

To test this hypothesis, we constructed scenes with multiple objects using the same concatenation method as in OVE scene construction framework.
We created two cases, as illustrated in Figure~\ref{FIG:segmentation}: i) Loc+Shape: objects in the scene belong to the same class but differ in shape and location;
ii) Loc+Class: objects differ in class and location but share similar shapes.
The evaluation task is object partial segmentation, \ie, correctly segmenting parts of objects. The evaluation metric is mIoU$_\text{I}$(\%), the mean IoU across all instances. 
3D grounding evaluates model capacity for text-visual alignment (\S\ref{sec-eval-grounding}), while 3D segmentation assesses visual modal understanding (\S\ref{sec-eval-segment}). We evaluate both tasks to study how shape and class differences affect model performance.

We selected Uni3D \cite{zhang2023uni3d} and PointBERT \cite{yu2022point} for evaluation, as they represent state-of-the-art performance. Considering the data adaptation issue, where models may require fine-tuning to perform well on new tasks, we added extra fully connected layers to fine-tune existing models on the tasks we constructed. We used two types of data for fine-tuning: `sc' refers to scenes containing objects from the same class, while `dc' refers to scenes with objects from different classes.

The experiment results are shown in Figure~\ref{fig:eval_segment}. Compared to their performance on single objects, which is 85.6\% for PointBERT and 78.2\% for Uni3D, we observe that the presence of distractors slightly hinders object partial segmentation. Figure~\ref{FIG:segmentation} visualizes some segmentation results, highlighting some errors these models make when predicting the IoU of targets. For example, when locating tables that are distracted by skateboards, both PointBERT and Uni3D incorrectly identify the skateboard as part of the bottom of the table.
This suggests that while these 3D encoders can distinguish between similar objects, there is still room for improvement for more complex scenarios.

Comparing the performance across different datasets (`sc' v.s. `dc'), we found that models trained on scenes containing objects from the same class performed better than those fine-tuned on data with different classes. This insight suggests that incorporating more distractor cases in the training data may be beneficial for enhancing 3D grounding capabilities. 
More details on segmentation results across different categories and shapes can be found in our project page.
 
\section{Conclusion and Future Work}

OVE advances point cloud LLM evaluation. It incorporates objects with controlled variations in properties (e.g., class, color, shape) and spatial relationships into real-world scanned scenes, enabling scene-level challenge customization. Besides, we develop an LLM-VLM-cooperated annotator to obtain fine-grained annotations for 3D grounding. 
Previously, evaluating which aspects a 3D model relies on for object differentiation was challenging; with OVE, such fine-grained evaluations are now more accessible. 
Our evaluation shows existing 3D models are limited in pure spatial reasoning when visual features (\eg, shape) are absent, which offer insights for improving 3D models, \eg, rethink position encoding effectiveness.


In the future, we plan to apply OVE to a broader range of scenes, including synthetic ones~\citep{jia2025sceneverse,yang20243d}, and expand our spatial predicate set to capture more spatial relationships. Besides, most object-level datasets only retain point information and lack mesh data, which means OVE cannot render 2D images with rich texture details, unlike real-scanned scenes in ScanNet. This prevents us from objectively evaluating 3D models that heavily rely on 2D features \cite{chen2024ll3da}. We plan to incorporate generative models to generate mesh data from point information to address this issue.

\section*{Acknowledgments}
We sincerely thank the anonymous reviewers for their valuable comments and suggestions, which have been instrumental in improving and refining this paper. 
We are also deeply grateful to our friends and families for their continuous encouragement throughout this research. 

This work is supported by the Guangdong Provincial Key Lab of Integrated Communication, Sensing and Computation for Ubiquitous Internet of Things (No.2023B1212010007).

\bibliography{aaai25}

\end{document}